\documentclass[10pt,twocolumn,letterpaper]{article}

\usepackage[pagenumbers]{cvpr} % To force page numbers, e.g. for an arXiv version

\usepackage{graphicx}
\usepackage{amsmath}
\usepackage{amssymb}
\usepackage{booktabs}
\usepackage{times}
\usepackage{epsfig}
\usepackage{color}
\usepackage{multirow}
\usepackage{makecell}
\usepackage{appendix}
\newcommand{\link}[1]{{\href{#1}{#1}}}

\usepackage[pagebackref,breaklinks,colorlinks]{hyperref}

\usepackage[capitalize]{cleveref}
\crefname{section}{Sec.}{Secs.}
\Crefname{section}{Section}{Sections}
\Crefname{table}{Table}{Tables}
\crefname{table}{Tab.}{Tabs.}

\def\cvprPaperID{5703} % *** Enter the CVPR Paper ID here
\def\confName{CVPR}
\def\confYear{2022}

\begin{document}

%%%%%%%%% TITLE - PLEASE UPDATE
% \title{\LaTeX\ Author Guidelines for \confName~Proceedings}
\title{Dynamic MLP for Fine-Grained Image Classification \\ by Leveraging Geographical and Temporal Information}

\iffalse
\author{First Author\\
Institution1\\
Institution1 address\\
{\tt\small firstauthor@i1.org}
% For a paper whose authors are all at the same institution,
% omit the following lines up until the closing ``}''.
% Additional authors and addresses can be added with ``\and'',
% just like the second author.
% To save space, use either the email address or home page, not both
\and
Second Author\\
Institution2\\
First line of institution2 address\\
{\tt\small secondauthor@i2.org}
}
\fi

\author{Lingfeng Yang$^{1 \dag}$, Xiang Li$^{1 *}$, Renjie Song$^{2}$, Borui Zhao$^{2}$, Juntian Tao$^{1}$, \\
Shihao Zhou$^{2}$, Jiajun Liang$^{2}$, Jian Yang$^{1}$\thanks{Corresponding author. $^\dag$Works is done as interns in Megvii Research. Lingfeng Yang, Xiang Li, Juntian Tao, and Jian Yang are from PCA Lab, Key Lab of Intelligent Perception and Systems for High-Dimensional Information of Ministry of Education, and Jiangsu Key Lab of Image and Video Understanding for Social Security, School of Computer Science and Engineering, Nanjing University of Science and Technology.} \\
$^{1}$Nanjing University of Science and Technology, $^{2}$Megvii Technology \\
{\tt\small \{yanglfnjust, xiang.li.implus, taojuntian, csjyang\}@njust.edu.cn} \\
{\tt\small \{songrenjie, zhoushihao, liangjiajun\}@megvii.com, zhaoborui.gm@gmail.com}
}
%  - yanglfnjust@njust.edu.cn (Primary)
%  - xiang.li.implus@njust.edu.cn 
%  - songrenjie@megvii.com 
%  - zhaoborui.gm@gmail.com 
%  - taojuntian@njust.edu.cn 
%  - zhoushihao@megvii.com 
%  - liangjiajun@megvii.com 
%  - csjyang@njust.edu.cn 

\maketitle

%%%%%%%%% ABSTRACT
\begin{abstract}
Fine-grained image classification is a challenging computer vision task where various species share similar visual appearances, resulting in misclassification if merely based on visual clues. Therefore, it is helpful to leverage additional information, e.g., the locations and dates for data shooting, which can be easily accessible but rarely exploited. In this paper, we first demonstrate that existing multimodal methods fuse multiple features only on a single dimension, which essentially has insufficient help in feature discrimination. To fully explore the potential of multimodal information, we propose a dynamic MLP on top of the image representation, which interacts with multimodal features at a higher and broader dimension. The dynamic MLP is an efficient structure parameterized by the learned embeddings of variable locations and dates. It can be regarded as an adaptive nonlinear projection for generating more discriminative image representations in visual tasks. To our best knowledge, it is the first attempt to explore the idea of dynamic networks to exploit multimodal information in fine-grained image classification tasks. Extensive experiments demonstrate the effectiveness of our method. The t-SNE algorithm visually indicates that our technique improves the recognizability of image representations that are visually similar but with different categories. Furthermore, among published works across multiple fine-grained datasets, dynamic MLP consistently achieves SOTA results\footnote{\link{https://paperswithcode.com/dataset/inaturalist}} and takes third place in the iNaturalist challenge at FGVC8\footnote{\link{https://www.kaggle.com/c/inaturalist-2021/leaderboard}}. 
Code is available at \link{https://github.com/ylingfeng/DynamicMLP.git}.
% \\xxx xxx xxx
\end{abstract}

\vspace{-20pt}
%%%%%%%%% BODY TEXT
\begin{figure}[t]
	\vspace{0pt}
	\begin{center}
		\setlength{\fboxrule}{0pt}
		\vbox{\includegraphics[width=\columnwidth]{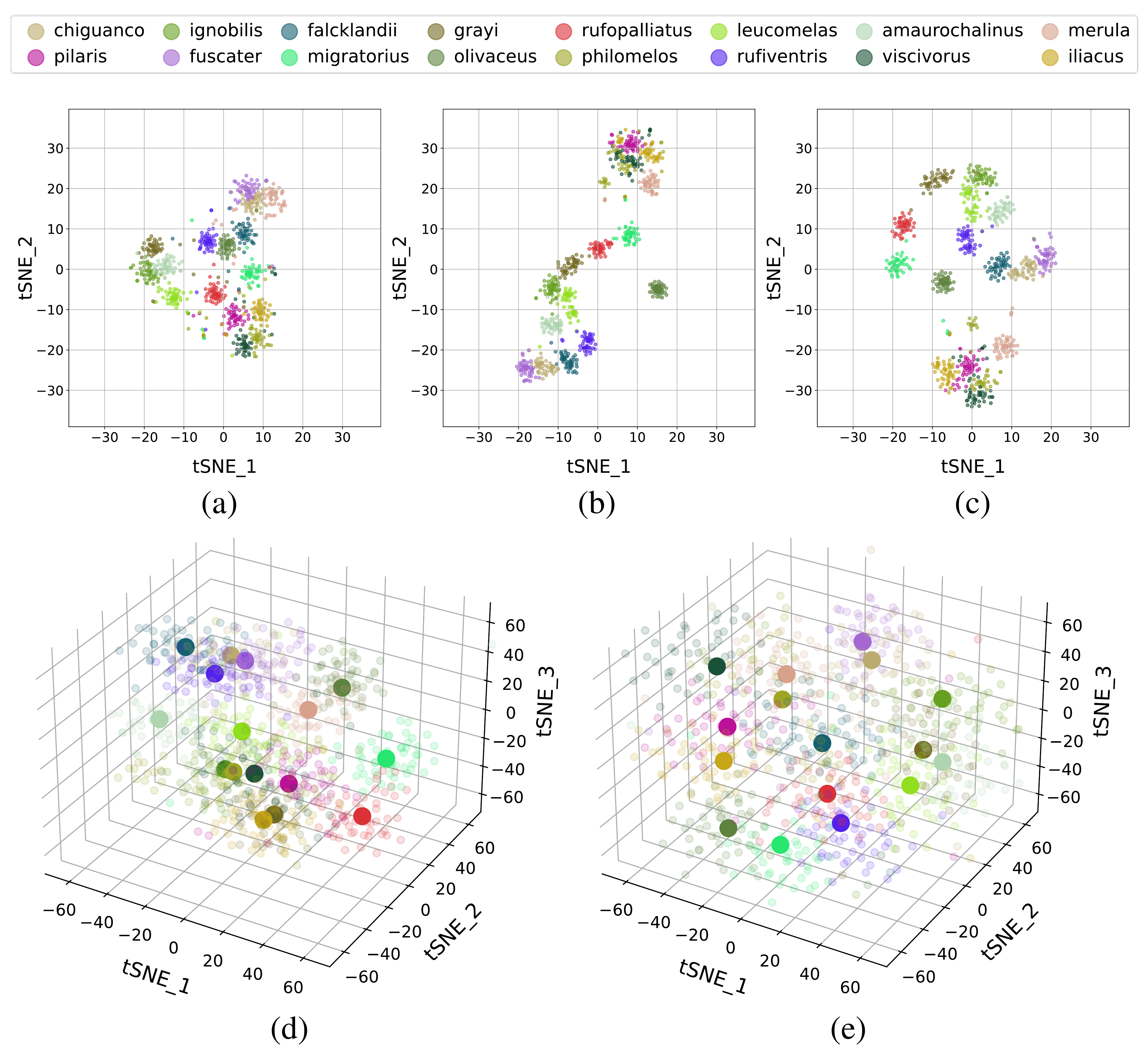}}
	\end{center}	
	\vspace{-18pt}
	\caption{
        Visualization of t-SNE~\cite{van2008visualizing} representations under well trained models. Various similar species from the genus Turdus in the iNaturalist 2021 dataset are depicted in different colors.
        (a): The visualization of an image-only model.
        (b): Concatenating the image, location, and date features before the classification head is a typical strategy of utilizing additional information to help classification. The concatenated representation is more discriminative than the original. The concatenation strategy can be regarded as a baseline for all methods that involve additional information.
        (c): Intuitively, our proposed dynamic MLP expands the diversity among different fine-grained species compared to the image-only or concatenation framework.
        (d): The 3-d visualization of image representation from the concatenation strategy.
        (e): The 3-d visualization of our dynamic MLP.
        }
	\label{fig:tsne}
	\vspace{-8pt}
\end{figure}
\section{Introduction}

\begin{figure}[t]
	\vspace{0pt}
	\begin{center}
		\setlength{\fboxrule}{0pt}
		\vbox{\includegraphics[width=\columnwidth]{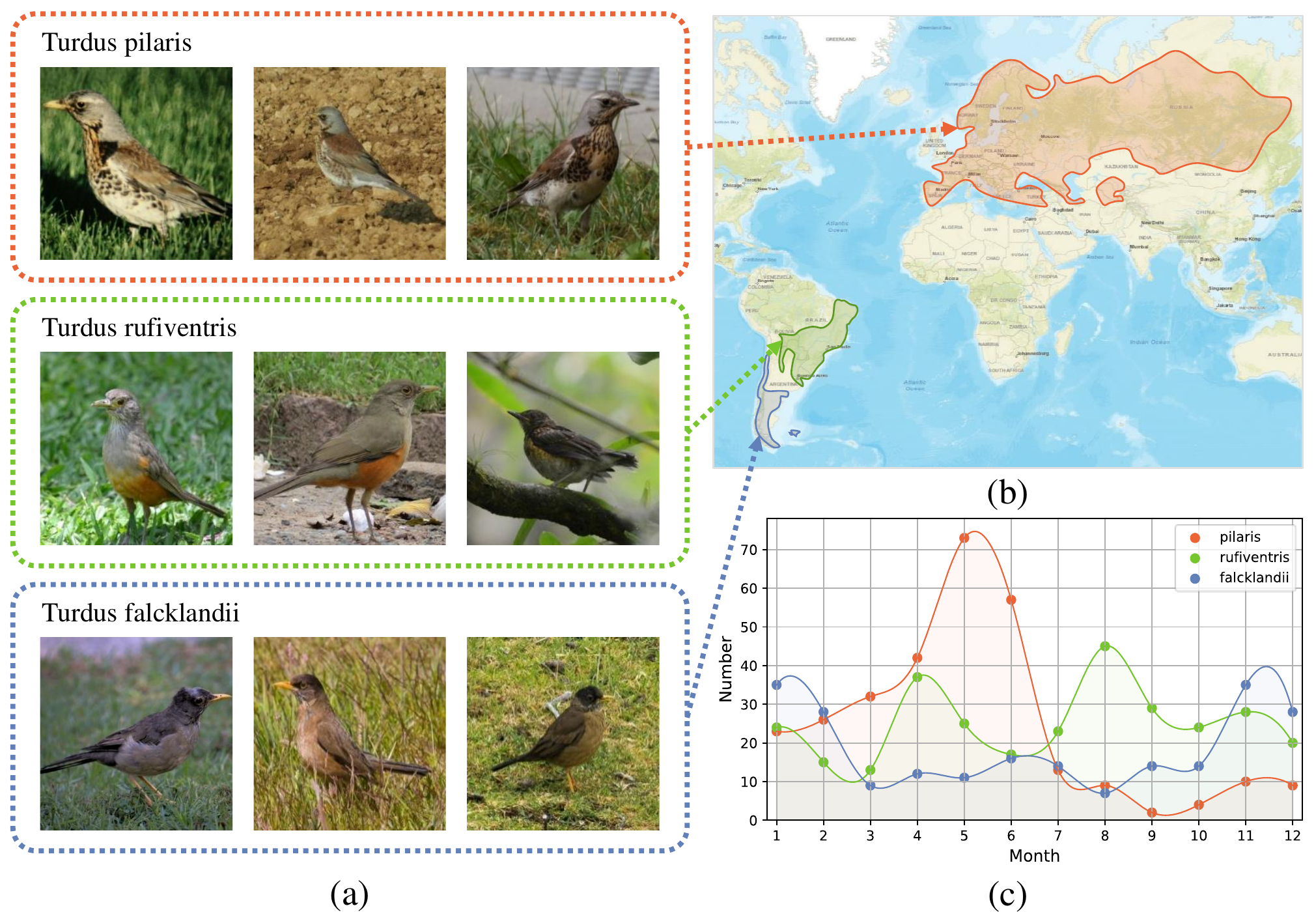}}
	\end{center}	
	\vspace{-10pt}
	\caption{
    The pilaris, rufiventris, and falcklandii are species belonging to the genus Turdus. (a): They are visually similar and hard to recognize based on their appearance. (b): Their habitats vary widely. (c): Their activity frequency varies throughout the year, so the data amount at different times is different.
	}
	\label{fig:geographical_temporal}
	\vspace{-8pt}
\end{figure}

Fine-grained image classification~\cite{zhang2014part,akata2015evaluation,xiao2015application,yang2018learning,chang2020devil,zhang2021multi,behera2021context} is a challenging computer vision task that distinguishes fine categories of objects or species. In contrast to traditional image classification~\cite{chen2015mxnet,he2016deep,huang2017densely,xie2017aggregated,li2019selective}, fine-grained image classification has difficulty identifying different species but with almost the same appearances. Those visually similar samples are practically impossible to differentiate if merely based on images.
Apart from several popular fine-grained methods that focus on the discriminative regions of images~\cite{zhang2014part,xiao2015application,yang2018learning,zhang2021multi,behera2021context},  multi-branch learning~\cite{zheng2019learning,chang2020devil,akata2015evaluation,gao2020channel,touvron2021grafit}, or particular data augmentations~\cite{li2020attribute,huang2020snapmix}, another available direction is to introduce additional information, e.g., geographical and temporal information, to help fine-grained image classification~\cite{tang2015improving,wittich2018recommending,chu2019geo,mac2019presence}.

% Clues other than visual information, e.g., locations and dates, can help distinguish species with similar appearances in fine-grained classification.
Images usually contain additional information, e.g., geographic location and time, denoting where and when to shoot, which can help with fine-grained classification.
For example, we choose three species from the genus Turdus, such as pilaris, rufiventris, and falcklandii, that are visually similar and thus difficult to distinguish (Fig.~\ref{fig:geographical_temporal} (a)). However, their living locations (Fig.~\ref{fig:geographical_temporal} (b)) and temporal distribution (Fig.~\ref{fig:geographical_temporal} (c)) are quite different.
It indicates that geographical and temporal information can be helpful to facilitate their accurate classification.
Further, additional information such as locations or dates has already been provided in some well-known datasets like BirdSnap~\cite{berg2014birdsnap}, PlantCLEF~\cite{goeau2016plant}, YFCC100M~\cite{thomee2016yfcc100m}, and iNaturalist~\cite{van2018inaturalist,van2021benchmarking},
and is widely available on the internet~\cite{gbif2001,flickr2004,imageclef2006,inaturalist2008}.

\begin{figure*}[t]
    \vspace{0pt}
	\begin{center}
		\setlength{\fboxrule}{0pt}
		\vbox{\includegraphics[width=\textwidth]{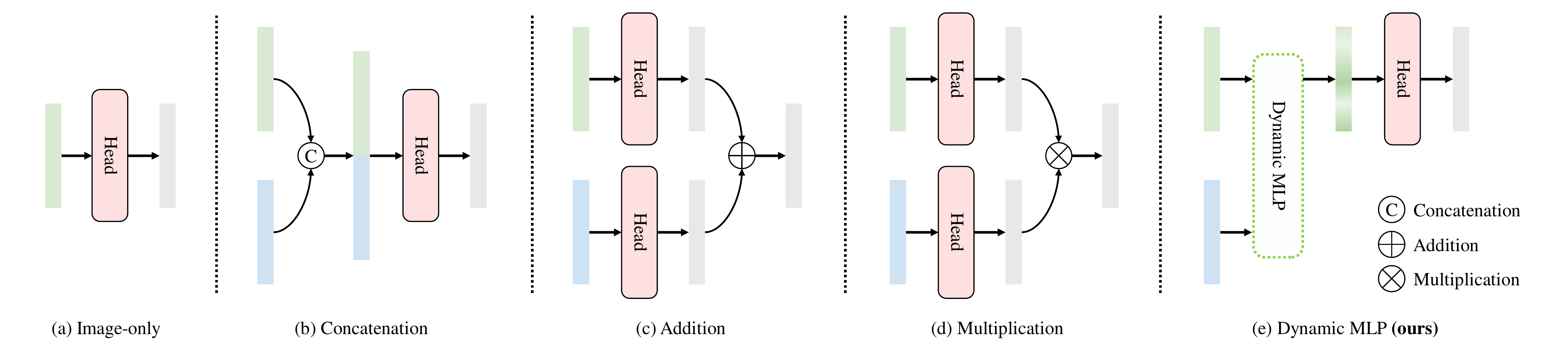}}
	\end{center}
	\vspace{-20pt}
	\caption{
    Comparisons of existing works with our dynamic MLP. The green and blue rectangles denote the image feature and multimodal feature, respectively. 
    (a) Image-only: Get predictions based solely on images~\cite{he2016deep,gao2019res2net,li2019selective}.
    (b) Concatenation: Concatenate the image features with the multimodal features in a channel-by-channel fashion~\cite{tang2015improving,terry2020thinking,salem2020learning,mai2020multi}.
    (c) Addition: Add both predictions from the last layer of the image and extra information for a joint prediction~\cite{chu2019geo}. 
    (d) Multiplication: Multiply both predictions~\cite{mac2019presence,terry2020thinking}.
    (e) Dynamic MLP: Features are fused through the dynamic MLP.
	}
	\label{fig:existing_works}
	\vspace{-6pt}
\end{figure*}

% There are several previous works that utilize the additional information in fine-grained image classification.
% \cite{tang2015improving} concatenates features extracted from the images and additional information to jointly produce the predictions (Fig.~\ref{fig:existing_works} (b)). However, this strategy does not reflect their internal interaction.
% \cite{wittich2018recommending} maintains a database that associates images with their additional information but with high memory consumption.
% \cite{chu2019geo} develops a post-processing network to refine the original prediction by adding the distribution derived from extra information (Fig.~\ref{fig:existing_works} (c)), 
% %which requires supernumerary training.
% which inevitably requires extra training cost.
% \cite{mac2019presence} trains two networks based on images and additional information separately and ensemble their predictions during inference (Fig.~\ref{fig:existing_works} (d)), which is prone to fall into a local optimum.
Several works have proposed the use of additional information in fine-grained image classification. \cite{tang2015improving,minetto2019hydra,terry2020thinking,salem2020learning,mai2020multi} directly concatenate the image feature with the multimodal feature before the final classification head. Furthermore, the addition~\cite{chu2019geo} and multiplication~\cite{mac2019presence,terry2020thinking} operations are adopted to fuse the features or predictions from the last network layer. The core strategies are summarized in Fig.~\ref{fig:existing_works} (b)-(d), respectively.
However, they only refer to a single dimension between image and multimodal features. As shown in Fig.~\ref{fig:tsne} (a)-(b), the concatenation strategy only pulls the cluster apart vertically—in one dimension. More analysis can be found in Sec.\ref{sec:interact_dimensions}.
% Cross attention~\cite{vaswani2017attention} are widely adopted in multimodal fusion field~\cite{yang2016stacked,arevalo2017gated,margffoy2018dynamic,gao2019dynamic,prakash2021multi}. However, they all deal with multimodal features between spatial feature maps~\cite{prakash2021multi}, or queries of textual embeddings~\cite{yang2016stacked,margffoy2018dynamic}, which cannot not be directly applied to multimodal fine-grained classification.

To fully exploit the potential effect of additional information, we propose to involve the higher-dimensional interaction between the multimodal representations.
Evidently, since species with similar appearances have related image features extracted by the same network, a fixed projection fails to distinguish accurately if their categories are different.
Especially when their locations are numerically close, existing multimodal methods technically lack the potential to make a distinction.
Thus, a dynamic, instance-wise projection, which maps similar image features to different positions in the feature space, can manage to classify accurately.
Different from existing works, we propose dynamic MLP to exploit the additional information in the form of adaptive perceptron weight to enhance the representation ability of image features (Fig.~\ref{fig:existing_works} (e)).
Specifically, the weights of dynamic MLP are generated from the multimodal features extracted from the additional information.
Then the image feature is updated by the dynamic MLP, where it is conditionally transformed by the adaptive weight.
The projecting process in the dynamic MLP involves high-dimensional interaction between the image feature and multimodal feature and is verified to be more efficient in separating the decision boundaries of similar species. In Fig.~\ref{fig:tsne}, the clusters denoting similar categories are pulled apart in all directions evenly, which is more effective than the former work.
%Further inspired by Sparse R-CNN~\cite{sun2020sparse} and CondInst~\cite{tian2020conditional}, we propose dynamic MLP for multimodal fine-grained image classification which adaptively adjust the weight of MLP based on additional information and project similar image features to discriminate representations.
% 不同于以往的方法，我们的动态MLP利用额外信息生成的特征作为图像前向传播的向导。
% 我们会动态的生成MLP的权重，去变换图像的表征。
% 我们的动机来自于：非常相似的图像，原始的image的feature也是相似的，在一个固定的投影下肯定也是相似的，所以我们要引入一个动态的投影过程，从而投影到不同的特征整空间的不同的位置。 

% we conduct extensive experiments on
% extensive experiments are conducted on four well-known benchmark datasets
To verify the effectiveness of our proposed dynamic MLP, we conduct extensive experiments on four well-known fine-grained image classification datasets (iNaturalist 2017, 2018, 2021~\cite{van2018inaturalist,van2021benchmarking} and YFCC100M-GEO100~\cite{thomee2016yfcc100m,tang2015improving}).
Notably, our dynamic MLP consistently outperforms previous works by 0.2\% $\sim$ 5.8\% top-1 accuracy across a variety of popular benchmarks.
% consistently achieves SOTA results on multiple fine-grained datasets
% We outperform the baseline which only uses images by 5.2\% top-1 accuracy on iNaturalist 2021. 
% We also improve the top-1 accuracy by range of 0.2\% to 5.8\% compared to other methods which also utilize additional information.
% Based on the proposed method, we achieve 3rd place in iNaturalist 2021 fine-grained dataset challenge.
% Notably, our method achieves 3rd place on iNaturalist challenge at FGVC8.

Overall, our contributions can be summarized as follows:

$\bullet$ We propose the dynamic MLP, an end-to-end trainable framework that jointly exploits images and additional information with high efficiency.
To the best of our knowledge, we are the first to use dynamic MLP in multimodal fine-grained image classification tasks.

$\bullet$ Compared to existing published works, our method consistently achieves SOTA results on multiple datasets, specifically 76.81\%, 83.67\%, and 91.39\% top-1 accuracy on iNaturalist 2017, 2018, and 2021, respectively. % among published works.

$\bullet$ An ensemble of dynamic MLP reaches 94.75\% top-1 accuracy on the iNaturalist 2021 dataset, which achieves 3rd place in the FGVC8~\cite{fgvc82021} at CVPR 2021.
% \\xxx xxx xxx 

%%%%%%%%%%%%%%%%%%%%%%%%%%%%%%%%%%%%%%%%%%%%%%%%%%%%%%%%%%%%%%%%%%%%%%%%%
\section{Related Work}
In this section, we review existing works on traditional fine-grained image classification that only deal with images. Then we summarize in fine-grained works that incorporate additional information. Finally, we introduce the development of the dynamic filter.
% such as the concatenation, logits combination, and knowledge graph.

\noindent\textbf{Fine-Grained Image Classification}:
To improve fine-grained image classification, several works have been proposed.
\cite{zhang2014part,xiao2015application,yang2018learning,zhang2021multi,behera2021context} detect the discriminative regions of an image to explore subtle details.
SnapMix~\cite{huang2020snapmix} uses the class activation map (CAM)~\cite{zhou2016learning} to reduce the label noise in fine-grained data augmenting. Similarly, Attribute Mix~\cite{li2020attribute} focuses on semantically meaningful attribute features from two images to identify the same super-categories.
FixRes~\cite{touvron2019fixing} investigates data augmentation and resolution to improve the classification performance.
Other researches focus on extracting more useful features from multi-channel networks~\cite{zheng2019learning,chang2020devil} or contrastive learning~\cite{akata2015evaluation,gao2020channel}.
TransFG~\cite{he2021transfg} and TPSKG~\cite{liu2021transformer} have recently used the Transformer architecture to improve classification performance.
The fine-grained methods generally suffer from a complex pipeline and enormous manual design. Moreover, most existing methods depend on large training resolution to perform effectively, which is incompatible with large datasets like iNaturalist. Our dynamic MLP can operate efficiently on the most basic baseline without particular training settings.

\noindent\textbf{Methods using Additional Information}:
Besides visual information, researchers have included additional information to improve classification performance.
Many existing works~\cite{tang2015improving,minetto2019hydra,terry2020thinking,salem2020learning,mai2020multi} combine the image feature with the additional multimodal feature directly through channel-wise concatenation.
%and then jointly pass them
% to pass through a unified network or classifier.
\cite{tang2015improving} is the first to introduce multimodal features, e.g., images, ages, and dates, extracted from MLP backbone network by concatenating them together to make a joint prediction.
Later, \cite{minetto2019hydra} introduces metadata to the geospatial land classification task, and \cite{salem2020learning} integrates dense overhead imagery with location and date into a general framework by concatenating the outputs of the context network.
% Works above may differ in their feature extraction strategies but share the same center point of multimodal fusion through feature concatenation.
%However, concatenating embeddings does not refine the image representation itself, leading to inaccurate classification when both location and date are close for two species with same appearance.
%On the contrary,
% Different from the concatenation methods where the projection matrix is shared,
% our proposed dynamic MLP can perform intrinsic, instance-wise transformation to the image representation guided by the additional information.
The concatenation methods share the projection weight for all samples, while our proposed dynamic MLP can perform an intrinsic, instance-wise transform to the image representation guided by additional information.
% \cite{tang2015improving}实验了多种图像与位置特征concat的网络宽度的对比试验，并将它们合并到一个CNN模型。
% \cite{terry2020thinking}
% \cite{salem2020learning}采用的是直接将两者各自通过encoder之后的特征相加的方式。
% \comment{简单的方式没法建模复杂的关系}
% \noindent\textbf{Logits Combination}:
Another fusion strategy aims to combine the output predictions of images and metadata through multiplication or addition operations.
\cite{mac2019presence} extracts location features by MLP to produce a prior distribution for fine-tuning the original predictions.
In GeoNet~\cite{chu2019geo}, the geolocation priors, post-processing models, and feature modulation models are utilized to leverage the additional information.
\cite{terry2020thinking} ensembles the result by multiplying the relative categorization probabilities.
%  image's and metadata's relative categorization probabilities.
% These methods often handle images and additional information separately, 
%which is hard for the network to obtain the optimal solution. 
% which is prone to fall into a local optimum.
% In contrast, our method is end-to-end trainable and can potentially learn the most discriminative representations for classification through joint backpropagation.
These methods often handle images and additional information separately, which could easily fall into a local optimum. In contrast, our framework is end-to-end trainable. Therefore, it could learn the most discriminative representations for classification through joint backpropagation.
% \cite{mac2019presence}提出了一个有效获取时空先验的方法，根据地理位置和时间，估计每个对象类别出现的概率，作为调整图像网络输出分布的辅助权重。
% \cite{terry2020thinking}归纳了concat模型与直接分布相乘的方法，使用更多的上下文本信息如温度位置等等显著提高自动识别的准确性。
% \comment{简单的方式没法建模复杂的关系}
% \noindent\textbf{Knowledge Graph}:
Using a knowledge graph is another way to handle additional information. \cite{wittich2018recommending} recommends a list including the most likely categories related to metadata based on a geo-aware database. \cite{nitta2020constructing} constructs a geospatial concept graph that contains the prior information for realizing the geo-aware classification.
% \cite{wittich2018recommending}维护了一个给定的地理位置和时间最有可能观察到的植物类别列表。此外还通过增加采样半径来聚合地理位置信息，改进聚合策略，并深度研究了数据离散化粒度对推荐质量和计算效率的影响。
% \cite{nitta2020constructing}通过使用在不同地方拍摄的大量图像和文本标签描述(从Flickr中收集)，将图像与其下属地理空间信息构建为由节点组成的知识图谱。并用图谱的知识为图像类别分布加权。
Since these methods are attached to a complex geographical or temporal knowledge graph, they often consume enormous memory costs and require specific manual design.

\begin{figure*}[t]
    \vspace{0pt}
	\begin{center}
		\setlength{\fboxrule}{0pt}
		\vbox{\includegraphics[width=0.98\textwidth]{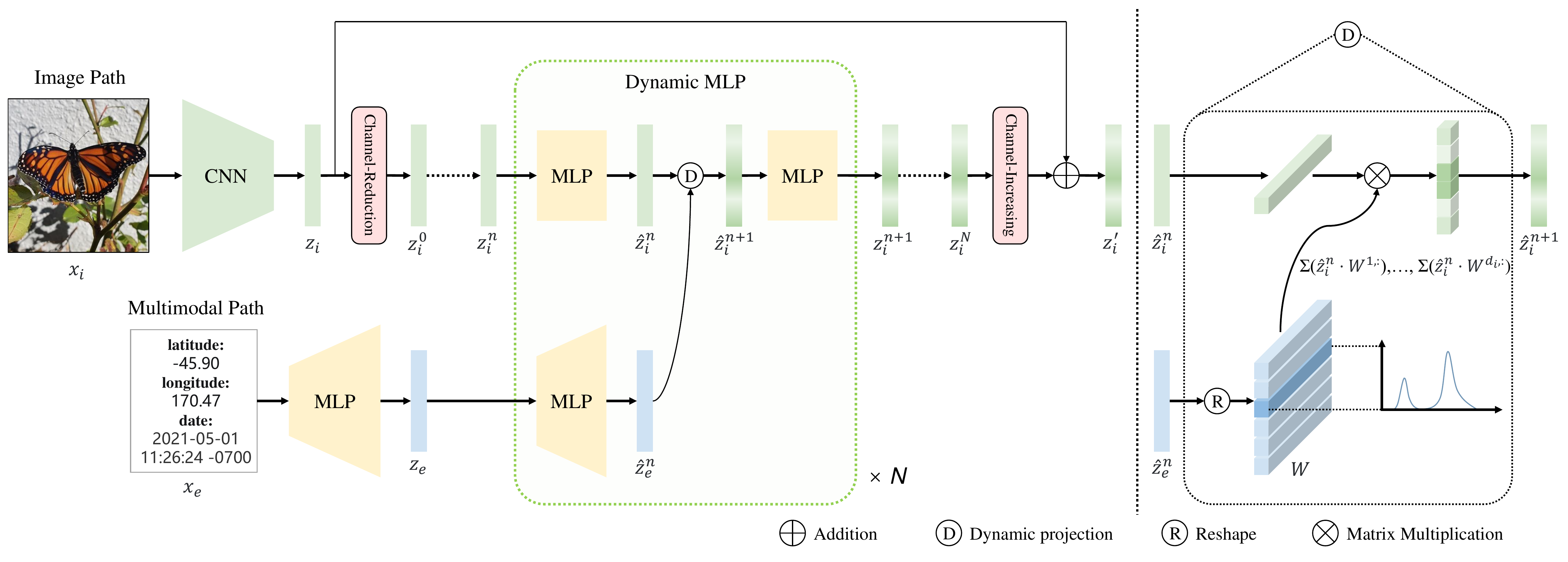}}
	\end{center}
	\vspace{-15pt}
	\caption{
    An overview of our proposed framework. It contains two paths for image and relational information, respectively. The dynamic MLP (in this figure a variant B is depicted) fuses two paths and produces the enhanced image representation.
    % After feeding forward through the backbone, the image feature is passed through dynamic MLP, where it is transformed by parameters generated adaptively based on the multimodal feature.
    The core of dynamic MLP is a dynamic projection, indicated as ``D'', which transforms the image representation by parameters adaptively generated from the multimodal feature.
    The dynamic MLP is employed recursively, where ``$N$'' denotes the stage number.
	}
	\label{fig:main_framework}
	\vspace{-6pt}
\end{figure*}

\noindent\textbf{Dynamic Filters}:
%Different from traditional convolution layers with shared filters, 
Dynamic filters are adaptively modulated based on the input features, showing impressive performance across various vision tasks.
CondConv~\cite{yang2019condconv}, DyNet~\cite{zhang2020dynet}, and Dynamic Conv~\cite{chen2020dynamic} learn dynamic parameters for multiple selective kernels.
\cite{jia2016dynamic,ha2016hypernetworks} dynamically generates the filters, in which the weights are conditioned on the input images.
Recently, Sparse R-CNN~\cite{sun2020sparse}, SOLOv2~\cite{wang2020solov2}, and CondInst~\cite{tian2020conditional} extend the idea to object detection and instance segmentation, and {\cite{yang2016stacked,margffoy2018dynamic,perez2018film} explore the usage in vision-language multimodal tasks.} 
In general, existing works derive dynamic filters from a single source, such as images. 
Besides, they all deal with multimodal features between spatial feature maps~\cite{prakash2021multi} or queries of textual embeddings~\cite{yang2016stacked,margffoy2018dynamic,perez2018film}, which cannot be directly applied to multimodal fine-grained classification.
Different from previous methods, our method introduces external information, i.e., locations and dates.
To the best of our knowledge, we are the first to use dynamic projection in fine-grained image classification to fuse the additional information.
% \\xxx xxx xxx

% Existing methods mainly focus on metadata encoding and only involve elementary data fusion strategy such as feature concatenation or logits combination. 
% Different from previous methods, we introduce additional feature to generate weight for dynamic MLP to transform images representations conditionally, which can be inserted into an end-to-end trainable framework.
% 现有的方法基本都是简单的连接特征或者是在测试阶段调整概率分布。
% 而我们首次使用动态权重的双特征交互方式，让图像与地理位置特征更深层次的融合。

\begin{figure*}[t]
    \vspace{0pt}
	\begin{center}
		\setlength{\fboxrule}{0pt}
		\vbox{\includegraphics[width=0.96\textwidth]{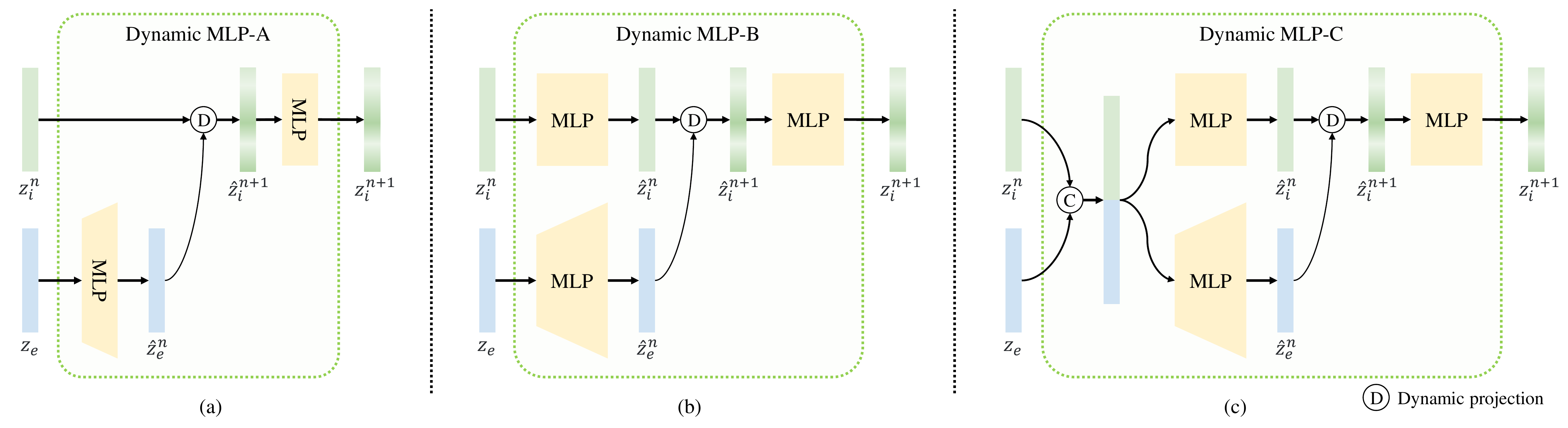}}
	\end{center}
	\vspace{-12pt}
	\caption{
	Comparisons between the three versions of dynamic MLP. (a) depicts the basic version and (b) expands the MLP depth. The image feature and multimodal feature in (a) and (b) are separately embedded, while they are concatenated together before projecting in (c).
	}
	\label{fig:dynamic_mlp}
	\vspace{-6pt}
\end{figure*}

%%%%%%%%%%%%%%%%%%%%%%%%%%%%%%%%%%%%%%%%%%%%%%%%%%%%%%%%%%%%%%%%%%%%%%%%%
\section{Method}
% In this section, a brief introduction to the overall framework is given
% This section introduces our proposed framework for fine-grained multimodal image classification.
In this section, we introduce a bilateral-path framework for multimodal fine-grained image classification that fuses additional information effectively. The framework contains an image path and a multimodal path, taking as input the image and additional information, respectively. First, the image feature is extracted by a CNN, whilst the multimodal feature is derived through the MLP backbone network. The adaptive projection with weighted multimodal features is then used to update it within a group of dynamic MLPs. The refined image representation produces the final prediction. Next, we elaborate on three dynamic MLP variants. Specifically, they are structures with basic implementation, with deeper embedding layers, and with concatenation inputs. Finally, we analyze the degree of interaction with multimodal information based on various methods.
% \\xxx xxx xxx
% 我们首先给出整体的端到端网络的流程介绍。随后我们详细的介绍我们改进的融合方法。
% 我们参考了\cite{mac2019presence}的残差全连接结构。

\subsection{Framework}\label{sec:framework}
% Inspired by existing works that utilize additional information, our overall
% \comment{先讲pipeline，大小维度都标一下，feature大小，每一个量的符号，
% encoding和embedding都直接在这里说了}
% architecture (Fig.~\ref{fig:main_framework}) for fine-grained image classification consists of two paths, i.e., the image path and multimodal path, to process the image and additional information, respectively.
% In our architecture (Fig.~\ref{fig:main_framework}), we design two paths, the image path and the multimodal path, for processing the image and additional information, respectively, inspired by existing works with additional information.
We design two paths in our architecture (Fig.~\ref{fig:main_framework}), the image path and the multimodal path, for processing the image and additional information, respectively, based on existing works with additional information.
Different from previous works, we propose a novel multimodal fusion method, termed ``dynamic MLP'', to refine and enhance the image representation based on the additional information.
We mark the original image feature after global average pooling as $\textbf{z}_{i}$ and the multimodal feature as $\textbf{z}_{e}$, providing the input image as $\textbf{x}_{i}$, and the additional information as $\textbf{x}_{e}$, respectively.
Our structure is designed as a recursive architecture since more dynamic projections can lead to better performance experimentally.
$\textbf{z}_{i}^{n}$ is defined in the following sections as the image representation updated $n$ times by the dynamic MLP, where $n \in \{1,2,...,N\}$.

% \begin{equation}
% % \setlength{\abovedisplayskip}{2pt}
% % \setlength{\belowdisplayskip}{2pt}
%     \textbf{z}_{i} = \mathcal{F}(\textbf{x}_{i}),
% \end{equation}
% where $\textbf{x}_{i}\in\mathbb{R}^{H\times W \times 3}$, $\textbf{z}_{i}\in\mathbb{R}^{D}$ denote the input image and output feature, respectively. 
% $\mathcal{F}(\cdot)$ denotes the function of CNN backbone.
Specifically, given the input image $\textbf{x}_{i}$, we can obtain its original embedding $\textbf{z}_{i}$ through the CNN backbone network, following the image path.
% At the same time, the multimodal path takes the additional information, e.g., latitude, longitude, date as input, and obtains multimodal feature $\textbf{z}_{e}$ by a residual MLP network.
Simultaneously, the multimodal path accepts additional information as input, such as latitude, longitude, and date, and obtains multimodal features $\textbf{z}_{e}$ via a residual MLP backbone network.
In detail, the additional information is first normalized to $[-1,1]$ and is then concatenated channel-wise:
\begin{equation}
\setlength{\abovedisplayskip}{2pt}
\setlength{\belowdisplayskip}{2pt}
    \hat{\textbf{x}}_{e} = \mathrm{Concat}(\{lat,lon,date\}),
\end{equation}
where $lat$, $lon$, and $date$ denote the latitude, longitude, and date related to an image, respectively.
$\mathrm{Concat}(\cdot)$ denotes the channel-wise concatenation and $\hat{\textbf{x}}_{e}\in\mathbb{R}^{3}$ denotes the intermediate encoding result of the additional information.
Then the additional information $\hat{\textbf{x}}_{e}$ is mapped to $\textbf{x}_{e}\in\mathbb{R}^{6}$:
% following the discriptions in PriorsNet~\cite{mac2019presence}:
\begin{equation}
\setlength{\abovedisplayskip}{2pt}
\setlength{\belowdisplayskip}{2pt}
    \textbf{x}_{e} = [\mathrm{Sin}(\pi \hat{\textbf{x}}_{e}), \mathrm{Cos}(\pi \hat{\textbf{x}}_{e})],
\end{equation}
where $\mathrm{Sin(\cdot)}$ and $\mathrm{Cos(\cdot)}$ denote the sine and cosine functions, respectively.
% considering their semantic periodicity
% Meanwhile we extract the multimodal feature by residual MLP network following the design in PriorsNet~\cite{mac2019presence}:
Finally, the multimodal feature $\textbf{z}_{e}$ is extracted by a residual MLP backbone network, following the descriptions in PriorsNet~\cite{mac2019presence}.
% \begin{equation}
%     \textbf{z}_{e} = \mathcal{G}(\textbf{x}_{e}),
% \end{equation}
% where $\mathcal{G}(\cdot)$ denotes the function of MLP backbone network and $\textbf{z}_{e}\in\mathbb{R}^{d_{e}}$ denotes the output multimodal feature.

% 下面一路处理额外信息，如地理位置、时间等。'E'代表编码过程，我们将额外信息归一化到[-$\pi$,$\pi$]的区间，并分别做正余弦映射到[-1,1]，目的在于约束输入的尺度，也符合时间与经度数值上周期变换的特点。
% 随后通过一个残差连接的MLP特征提取网络，得到额外信息特征，该特征为图像特征前向传播的MLP网络动态生成提供权重。
% 令额外信息输入为纬度$lat\in[-90,90]$，经度$lon\in[-180,180]$，时间$date\in[0,365]$，我们对它们进行归一化并做正余弦映射得到$x_{e}$，并通过残差MLP网络$\mathcal{G}(\cdot)$提取额外特征$z_{e}\in\mathbb{R}^{d}$。
% \begin{center}
%     $lat = lat/90$\\
%     $lon = lon/180$\\
%     $date = (date/365)\times 2-1$\\
%     $x_{e} = (lat,lon,date) \times \pi $
%     $x_{e} = (sin(x_{e}),cos(x_{e}))$\\
%     $z_{e} = \mathcal{G}(x_{e})$
% \end{center}

After the image and multimodal features are obtained, they are fused via the dynamic MLP.
% To fuse the multimodal features, we create the dynamic MLP whose weights are adaptively generated from the multimodal feature to transform the image feature. Notably, a skip connection is added to retain original image representation. %outward our proposed dynamic MLP.
% To limit model complexity
To save memory costs and runtime, the image feature is reduced to ${z}_{i}^{0}$ with lower dimensions by a channel-reduction layer.
% space RD to Rd where D > d: through a downsample layer 
% \begin{equation}
% \setlength{\abovedisplayskip}{2pt}
% \setlength{\belowdisplayskip}{2pt}
%     \textbf = f_{d}(\textbf{z}_{i}),
% \end{equation}
% where $f_{d}$ denotes the downsample layer function.
% Generally, the output dimension ($D$) of CNN models~\cite{he2016deep,szegedy2016rethinking} is fixed to 2048, whilst the reduced dimension ($d$) setting is discussed in Table~\ref{tab:dimension}.
The dynamic MLP takes $\textbf{z}_{i}^{0}$ and $\textbf{z}_{e}$ as the initial inputs, where the image feature is dynamically projected by the corresponding multimodal feature.
We obtain the enhanced image representation $\textbf{z}_{i}^{N}$ after $N$ recursive dynamic MLP blocks.
The details of dynamic MLP are elaborated in Sec.\ref{sec:dynamic_mlp_structures}.
Next, $\textbf{z}_{i}^{N}$ is expanded to align the shape with the original input $\textbf{z}_{i}$ by a channel-increasing layer.
Then we employ a skip connection between the adjusted and original features to produce the final predictions.
%generates a higher dimension feature from $\textbf{z}_{i}^{N}$, in the same size as the original image feature $\textbf{z}_{i}$, which is then added to the result through a skip connection for producing the final predictions:
% Then the ultimate prediction is obtained by:
% \begin{equation}
% \setlength{\abovedisplayskip}{2pt}
% \setlength{\belowdisplayskip}{2pt}
%     \hat{\textbf{y}} = h(f_{u}(\textbf{z}_{i}^{N}) +\textbf{z}_{i}),
% \end{equation}
% where $h$ denotes the linear classifier.
%where $f_{u}(\cdot)$ and $\hat{\textbf{y}}$ denote the up-sample layer function and the final prediction, respectively.
% 最后dynamic MLP的输出$z_{i}^{N}$经过一层全连接生成logits做分类。

\begin{figure}[t]
	\vspace{0pt}
	\begin{center}
		\setlength{\fboxrule}{0pt}
		\vbox{\includegraphics[width=\columnwidth]{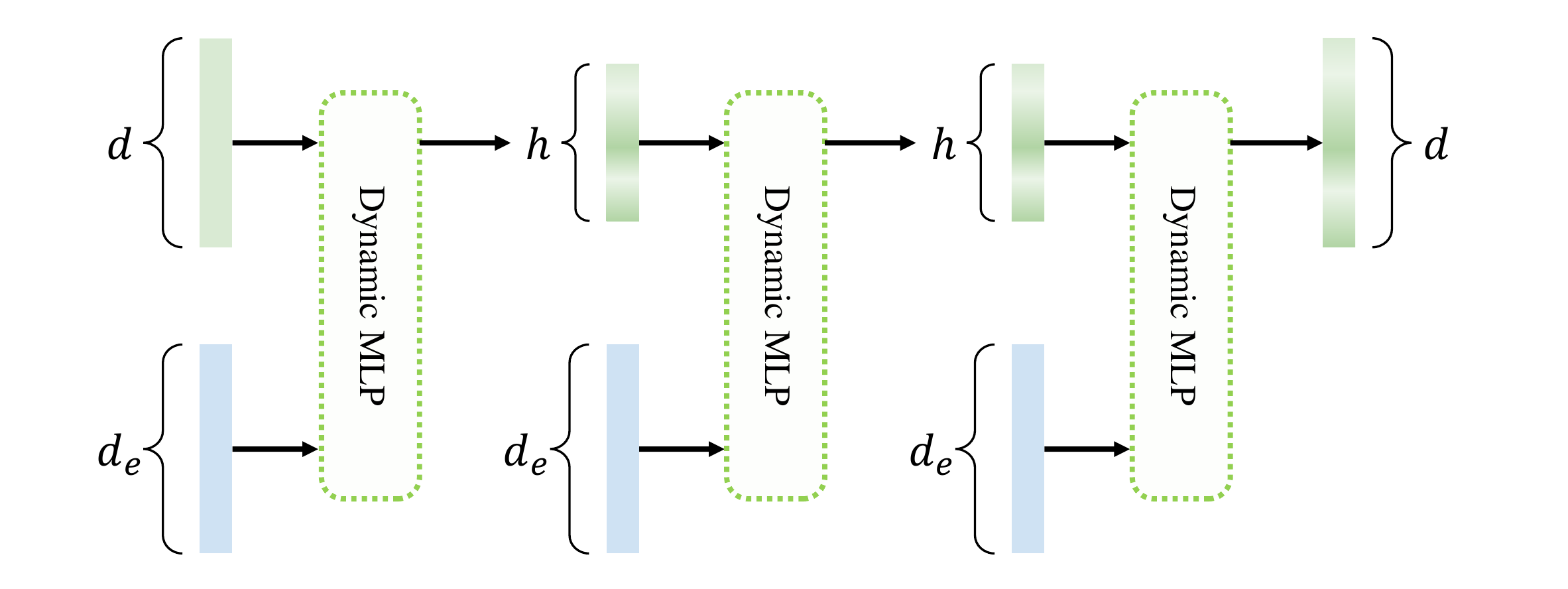}}
	\end{center}	
	\vspace{-12pt}
	\caption{
    The channel width in dynamic MLP, which resembles a bottleneck architecture ($N=3$).
    }
	\label{fig:stage_number}
	\vspace{-6pt}
\end{figure}

The whole framework not only learns to extract high-quality visual contextual clues but also tries to focus on optimal attention through image representations based on the guidance of multimodal features.
% , which is verified by the visualization (Fig.~\ref{fig:cam}) produced with CAM~\cite{zhou2016learning}.
% intuitively verify that it is better than existing works.
Different from the traditional convolution layer or linear layer, the parameters of which are fixed for all instances, the projector’s weights in our framework are dynamically generated and conditioned based on the instance-wise additional information. 
It eases the recognition difficulty if two different species are similar in appearance.
Essentially, our dynamic MLP fuses the multimodal features into a higher and broader dimensions than the single dimension produced by the former methods, which is further discussed in Sec.\ref{sec:interact_dimensions}.
Through end-to-end optimization, our network can learn more generalized and discriminative features, which significantly outperforms the baseline image-only network as well as surpasses existing published works that utilize additional information in ordinary ways.

\begin{table*}[t]
	\renewcommand\arraystretch{1.1}
	%\begin{center}
	%	\setlength{\fboxrule}{0pt}
	%	\input{tabs/table3.tex}
	%\end{center}
	\centering
	\footnotesize \setlength{\tabcolsep}{4.5pt}
		\resizebox{0.83\textwidth}{!}{
			\begin{tabular}
                {l|l|l|cc|cc|cc|cc}
                \hline
                % \multicolumn{3}{l|}{\textbf{Settings \textbackslash Datasets}} &
                % \multicolumn{2}{c|}{\textbf{YFCC-GEO}} & 
                % \multicolumn{2}{c|}{\textbf{iNat18}} & 
                % \multicolumn{2}{c|}{\textbf{iNat21 Mini}} & 
                % \multicolumn{2}{c}{\textbf{iNat21 Full}} \\
                % \hline
                % Backbone & Method & Reference & Top-1 & Top-5 & Top-1 & Top-5 & Top-1 & Top-5 & Top-1 & Top-5 \\
                Backbone & Method & Reference & 
                \multicolumn{2}{c|}{YFCC-GEO} & 
                \multicolumn{2}{c|}{iNat18} & 
                \multicolumn{2}{c|}{iNat21 Mini} & 
                \multicolumn{2}{c}{iNat21 Full} \\
                \hline\hline
                % \textit{\scriptsize{224×224}} 
                \multirow{6}{*}{ResNet-50}
                & Baseline & -- & 47.650 & 77.950 & 64.345 & 85.348 & 67.924 & 86.563 & 79.830 & 93.363 \\
                \cline{2-11}
                & ConcatNet~\cite{tang2015improving} & ICCV 2015 & 47.600 & 78.600 & 77.135 & 92.754 & 78.326 & 92.456 & 87.183 & 96.639 \\
                & PriorsNet~\cite{mac2019presence} & ICCV 2019 & 41.850 & 72.250 & 70.097 & 86.625 & 78.019 & 92.362 & 86.784 & 96.523 \\
                & GeoNet~\cite{chu2019geo} & ICCV 2019 & 48.800 & 80.000 & 75.035 & 91.169 & 77.738 & 91.875 & 86.727 & 96.229 \\
                & EnsembleNet~\cite{terry2020thinking} & MEE 2020 & 50.800 & 81.950 & 73.692 & 89.769 & 72.148 & 87.926 & \textbf{87.595} & 96.519 \\
                % \cline{2-11}
                & Dynamic MLP \textbf{(ours)} & -- & \textbf{53.200} & \textbf{83.850} & \textbf{78.220} & \textbf{93.433} & \textbf{78.751} & \textbf{92.896} & 87.283 & \textbf{96.749} \\
                \hline
                \multirow{6}{*}{ResNet-101} 
                & Baseline & -- & 52.650 & 83.050 & 66.953 & 87.440 & 70.710 & 88.477 & 82.080 & 94.639 \\
                \cline{2-11}
                & ConcatNet~\cite{tang2015improving} & ICCV 2015 & 52.800 & 84.050 & 78.576 & 93.585 & 80.564 & 93.582 & 88.494 & 97.135 \\
                & PriorsNet~\cite{mac2019presence} & ICCV 2019 & 47.650 & 77.150 & 72.521 & 87.632 & 80.111 & 93.433 & 88.439 & 97.135 \\
                & GeoNet~\cite{chu2019geo} & ICCV 2019 & 53.800 & 84.350 & 76.529 & 92.418 & 79.165 & 92.700 & 87.987 & 96.747 \\
                & EnsembleNet~\cite{terry2020thinking} & MEE 2020 & 52.550 & 84.400 & 76.934 & 91.501 & 75.295 & 89.775 & \textbf{88.800} & 96.941 \\
                % \cline{2-11}
                & Dynamic MLP \textbf{(ours)} & -- & \textbf{54.500} & \textbf{84.450} & \textbf{79.248} & \textbf{93.847} & \textbf{80.574} & \textbf{93.860} & 88.482 & \textbf{97.245} \\
                \hline
                \multirow{6}{*}{SK-Res2Net-101} 
                & Baseline & -- & 54.400 & 83.500 & 74.150 & 91.710 & 76.102 & 91.456 & 86.178 & 96.400 \\
                \cline{2-11}
                & ConcatNet~\cite{tang2015improving} & ICCV 2015 & 55.550 & 85.600 & 82.081 & 94.874 & 83.706 & 94.932 & 91.200 & 97.992 \\
                & PriorsNet~\cite{mac2019presence} & ICCV 2019 & 50.600 & 78.800 & 77.720 & 90.129 & 83.600 & 94.925 & 90.773 & 98.010 \\
                & GeoNet~\cite{chu2019geo} & ICCV 2019 & 56.050 & 84.950 & 78.912 & 93.740 & 81.360 & 93.712 & 90.070 & 97.602 \\
                & EnsembleNet~\cite{terry2020thinking} & MEE 2020 & 53.800 & 82.750 & 80.468 & 93.421 & 80.277 & 92.743 & 91.137 & 97.831 \\
                % \cline{2-11}
                & Dynamic MLP \textbf{(ours)} & -- & \textbf{56.800} & \textbf{85.950} & \textbf{83.673} & \textbf{95.636} & \textbf{84.694} & \textbf{95.337} & \textbf{91.397} & \textbf{98.213} \\
                \hline
			\end{tabular}
		}
	\vspace{-5pt}
	\caption{
    Comparisons to previous SOTA multimodal works on the YFCC100M-GEO100~\cite{tang2015improving} and the iNaturalist 2018, 2021~\cite{van2018inaturalist,van2021benchmarking} datasets.
    % We conduct experiments on ResNet-50~\cite{he2016deep}, ResNet-101~\cite{he2016deep} and SK-Res2Net-101~\cite{li2019selective,gao2019res2net} to demonstrate the effectiveness of our method.
    For each experiment, we present the top-1 (left) and top-5 (right) accuracy.
	}
	\vspace{-7pt}
	\label{tab:sota}
\end{table*}

\begin{table}[t]
    \vspace{5pt}
	%\scriptsize
	\small
	\centering
	\renewcommand\arraystretch{1.1}
	\newcommand{\tabincell}[2]{\begin{tabular}{@{}#1@{}}#2\end{tabular}}
	\resizebox{\columnwidth}{!}{
		\begin{tabular}
            {l|l|c|c|l|c}
            \hline
            Type & Method & Flops & \#Params & Res & Acc (\%) \\
            \hline\hline
            \multirow{3}{*}{Image-only}
            & Baseline                              & 8.9 G  & 56.8 M  & 224  & 71.0 \\
            & FixSENet~\cite{touvron2019FixRes}     & 20.8 G & 123.5 M & 224* & 75.4 \\% NeurIPS 2019
            & TransFG~\cite{he2021transfg}          & 49.2 G & 90.04 M & 304  & 71.7 \\% Arxiv 2021
            \hline
            \multirow{5}{*}{Multimodal}
            & ConcatNet~\cite{tang2015improving}    & 8.9 G  & 58.7 M  & 224  & 76.3 \\% ICCV 2015
            & PriorsNet~\cite{mac2019presence}      & 8.9 G  & 58.6 M  & 224  & 71.1 \\% ICCV 2019
            & GeoNet~\cite{chu2019geo}              & 8.9 G  & 58.6 M  & 224  & 74.0 \\% ICCV 2019
            & EnsembleNet~\cite{terry2020thinking}  & 8.9 G  & 58.6 M  & 224  & 73.8 \\% MEE 2020
            & Dynamic MLP \textbf{(ours)}           & 8.9 G  & 64.1 M  & 224  & \textbf{76.8} \\ 
            \hline
		\end{tabular}
	}
	\vspace{-5pt}
	\caption{
    We report the best-published result of a single model from state-of-the-art fine-grained image-only methods and multimodal methods on the iNaturalist 2017 dataset. ``Res'' denotes the input image resolution. *The test resolution in FixSENet~\cite{touvron2019FixRes} is larger than the training phase which brings extra gain in performance.
    }
	\vspace{-5pt}
	\label{tab:inat17}
\end{table}

\subsection{Dynamic MLP Structures}\label{sec:dynamic_mlp_structures}
% \comment{把batch size去掉，就直接单张图的流程。开头的时候回溯一下我们要用额外特征得到动态权重，用动态权重去MLP，forward图像特征。这个动态的权重可以用额外特征生成，也可以用concat的特征生成。
% 图5说明image path和extra path
% 画图直接画concat的算法把，把c->c->c*c，中间这个流程的交互就去掉。就按照最终的画
% 就提一嘴普通的方式了，然后上下两路都融合，或者某一路融合的对比实验，在表三里面}
In this section, we first describe the elementary implementation of our proposed dynamic MLP (Fig.~\ref{fig:dynamic_mlp} (a)). Then two of its improved variants (Fig.~\ref{fig:dynamic_mlp} (b)-(c)) are discussed.
% The overall architecture is shown in Fig.~\ref{fig:dynamic_mlp}.
% which enhance the feature's representation ability and interactivity, respectively.

\noindent\textbf{Dynamic MLP-A}:
Inspired by dynamic filters~\cite{jia2016dynamic,yang2019condconv,zhang2020dynet,chen2020dynamic,sun2020sparse}, we propose an iterable structure, namely dynamic MLP, which adaptively promotes the representation ability of the image feature guided by the multimodal feature.
In Fig.~\ref{fig:dynamic_mlp} (a), we show a single unit of dynamic MLP-A, the most concise implementation of dynamic MLP.
Each input image feature for the current dynamic MLP block is the output of the former block.
For the multimodal input for each block, we uniformly use the original feature.
Notably, all the recursive blocks of dynamic MLP are identical except for their channel dimensions.
% To achieve a lightweight structure, we design a bottleneck architecture (Fig.~\ref{fig:stage_number}) by setting smaller hidden channel ($h$) for the intermediate dynamic MLP blocks compared to the original input dimension ($d$). 
We create a bottleneck architecture (Fig.~\ref{fig:stage_number}) by specifying a smaller hidden channel ($h$) for the intermediate dynamic MLP blocks than the original input dimension ($d$).
Experimentally, the detailed channel settings are discussed in Table~\ref{tab:dimension}.
% only the input dimension of the first iteration and the output dimension of the last iteration in dynamic MLP to $d$, e.g. 256, same to the multimodal feature's dimension. 
% Besides, the intermediate feature dimension is set to a smaller number $h$, e.g. 64, denoting the hidden channel. 
% The structure resembles a bottleneck architecture Table~\ref{tab:dimension} 

Specifically, it takes as input the image and multimodal feature and outputs the projected image feature for one iteration.
First, the weight of the dynamic projector is generated based on the multimodal feature:
% it generates weights for through a single linear layer:
\begin{equation}
\setlength{\abovedisplayskip}{2pt}
\setlength{\belowdisplayskip}{2pt}
    \mathbf{W} = \mathrm{Reshape}(f(\textbf{z}_{e})),
\end{equation}
where $\mathrm{Reshape}(\cdot)$ reformulates a 1-d feature into a 2-d matrix, and $f(\cdot)$ denotes the fully connected layer.
The following MLP structure can be illustrated as:
% i.e., projected by the generated conditional weight:
% , followed by a normalization and ReLU.
% The weight generation function $g(\cdot)$ is specifically a fully connected layer followed by reshaping operation. Then the forward propagation process (Eq.~\eqref{equ:generate_weight} and Eq.~\eqref{equ:dynamic_fc}) can be rewrited as:
% Thus We can rewrite Eq.~\eqref{equ:generate_weight} and Eq.~\eqref{equ:dynamic_fc} as:
\begin{equation}
\setlength{\abovedisplayskip}{2pt}
\setlength{\belowdisplayskip}{2pt}
% \begin{split}
   \textbf{z}_{i}^{n+1} = \mathrm{ReLU}\big(\mathrm{LN}(f(\textbf{z}_{i}^{n};\mathbf{W}))\big),
                        % &= \delta(\mathcal{N}(\hat{\textbf{z}}_{i}^{n}\times \mathbf{W}^{T})),
% \end{split}
\end{equation}
where $\mathrm{ReLU}(\cdot)$ and $\mathrm{LN}(\cdot)$ denote ReLU activation function and layer normalization~\cite{ba2016layer}, respectively.
% Notably in dynamic MLP-A, $\hat{\textbf{z}}_{i}^{n}\in\mathbb{R}^{d_{i}}$ is identical to the image feature $\textbf{z}_{i}^{n}$.
Notably, the above process is iterated until $N$ recursions are completed.

\noindent\textbf{Dynamic MLP-B}: 
In Fig.~\ref{fig:dynamic_mlp} (b), we exploit a deeper-in-depth version of dynamic MLP-A by expanding MLP layers. The extended MLP blocks are simple sequences of regular fully connected layers, layer normalization~\cite{ba2016layer}, and ReLU activation function.
%\rewrite{The embedding MLP blocks in dynamic MLP-B (Fig.~\ref{fig:dynamic_mlp} (b)) are enlarged.
% The structure is further developed by deeper modeling of more linear layers, for a single layer may not be sufficient to characterize projection relationship. 
% More specifically, we additionally embed the image feature and expand the weight generation function:
% \begin{equation}\label{equ:embed_xi}
% % \setlength{\abovedisplayskip}{2pt}
% % \setlength{\belowdisplayskip}{2pt}
%     \hat{\textbf{z}}_{i}^{n} = f_{i}\big(\delta(f_{i}(\textbf{z}_{i}^{n}))\big),
% \end{equation}
% \begin{equation}\label{equ:embed_xe}
% % \setlength{\abovedisplayskip}{2pt}
% % \setlength{\belowdisplayskip}{2pt}
%     \hat{\textbf{z}}_{e} = f_{e}\big(\delta(f_{e}(\textbf{z}_{e}))\big),
% \end{equation}
% where $f_{i}$ and $f_{e}$ denote the fully connected layer for image path and multimodal path, respectively.
%Note that the expanded blocks in dynamic MLP-B are made of regular fully connected layers, layer normalization and ReLU activation function.
%For the MLP before and after the dynamic fully connected layer, we additionally construct a single fc-ln-relu block to refine the features.
% At the end of dynamic-MLP, we add a linear layer along with normalization and ReLU.
%Empirically, a deeper network~\cite{he2016deep}
% \rewrite{add a reference, e.g., resnet} 
%can enhance the representation ability of feature, 
%thus we expect dynamic MLP-B to achieve better performance on classification.}
% be sufficient to characterize projection relationship.

\noindent\textbf{Dynamic MLP-C}:
The inputs for the above structures are non-interactive until the dynamic projection. However, in dynamic MLP-C (Fig.~\ref{fig:dynamic_mlp} (c)), the inputs are concatenated together channel-wise before any of the embedding layers.
% \begin{equation}
% \setlength{\abovedisplayskip}{2pt}
% \setlength{\belowdisplayskip}{2pt}
%     \textbf{c}^{n} = \mathrm{Concat}(\{\textbf{z}_{i}^{n},\textbf{z}_{e}\}),
% \end{equation}
% where $\textbf{z}_{i}^{n}\in\mathbb{R}^{d_{i}}$, $\textbf{z}_{e}\in\mathbb{R}^{d_{e}}$, $\textbf{c}^{n}\in\mathbb{R}^{d_{i}+d_{e}}$ denote image feature, multimodal feature, and their concatenation, respectively.
Since the image feature and multimodal feature have mutual information supplement potential, we speculate that they can jointly generate better conditional weight to guide the transformation of image representation.
Additionally, we perform an ablating study on the situations where only one input is replaced by the concatenated feature while the other remains unchanged, which is discussed in Table~\ref{tab:structure}.

\subsection{Interactive Dimensions}\label{sec:interact_dimensions}
In this section, we compare the differences in interactive dimensions on multimodal features between previous works and our dynamic MLP. We aim to demonstrate that previous works have limitations in enlarging the distance of image representations in the feature space.
As for the concatenation strategy, given the image feature $\textbf{z}_{i} \in \mathbb{R}^{d_{i}}$ and multimodal feature $\textbf{z}_{e} \in \mathbb{R}^{d_{e}}$, where $d_{i}$ and $d_{e}$ are the feature channels, we can extend ${z}_{i}$ and ${z}_{e}$ to the ``$d_{i}+d_{e}$'' dimension by padding zeros:
$\textbf{z}'_{i} = \{\textbf{z}_{i}^{1},...,\textbf{z}_{i}^{d_{i}},0,...,0\}$, where $d_{e}$ zeros are padded after ${z}_{i}$. In the same way we can get $\textbf{z}'_{e}=\{0,...,0,\textbf{z}_{e}^{1},...,\textbf{z}_{e}^{d_{e}}\}$ with $d_{i}$ zeros.
Then the final predictions can be derived as:
\begin{equation}
\setlength{\abovedisplayskip}{2pt}
\setlength{\belowdisplayskip}{2pt}
    y = h(\mathrm{Concat}(\{\textbf{z}_{i},\textbf{z}_{e}\})) = h(\textbf{z}'_{i}+\textbf{z}'_{e}),
\end{equation}
where y denotes the prediction and $h(\cdot)$ the classifier head, respectively.
The fusion procedure of the addition strategy is exactly an element-wise addition between dual outputs:
\begin{equation}
\setlength{\abovedisplayskip}{2pt}
\setlength{\belowdisplayskip}{2pt}
    y = h(\textbf{z}_{i})+h(\textbf{z}_{e}).
\end{equation}
Similarly, the multiplication strategy can be depicted as:
\begin{equation}
\setlength{\abovedisplayskip}{2pt}
\setlength{\belowdisplayskip}{2pt}
    y = h(\textbf{z}_{i}) \cdot h(\textbf{z}_{e})=exp(log(h(\textbf{z}_{i}))+log(h(\textbf{z}_{e}))).
\end{equation}
The common characteristic of the previous fusion strategy is essentially to add the representation or predictions based on images, locations, and dates, which involves only one dimension within multimodal features.
Different from the former methods, the weights of dynamic MLP are adaptively generated from the location and date, and the dynamically refined image representation is obtained after an instance-wise projection between the original image representation and the weighted features of locations and dates:
\begin{equation}
\setlength{\abovedisplayskip}{2pt}
\setlength{\belowdisplayskip}{2pt}
    y=h(f(\textbf{z}_{i},\textbf{z}_{e}))=
    h(\Sigma{(\textbf{z}_{i}^{:}\cdot\textbf{z}_{e}^{:,1})},...,\Sigma{(\textbf{z}_{i}^{:}\cdot\textbf{z}_{e}^{:,d_{i}})}).
\end{equation}
The analysis indicates that our fusion strategy introduces a higher-dimensional interaction. As a result, it separates the cluster of image representations to a larger extent in all directions (Fig.~\ref{fig:tsne}).
% \\xxx xxx xxx

\iffalse
\begin{table}[t]
	%\scriptsize
	\small
	\centering
	\renewcommand\arraystretch{1.1}
	\newcommand{\tabincell}[2]{\begin{tabular}{@{}#1@{}}#2\end{tabular}}
	%\begin{subtable}{0.44\textwidth}
	\resizebox{0.7\columnwidth}{!}{
		\begin{tabular}
            {l|c|c|c|c}
            \hline
            Backbone & Res & MS$_{\text{test}}$ & Ten-crop & Acc (\%) \\
            \hline\hline
            SK-101 & 224 & & & 91.397 \\
            SK-101 & 224 & $\checkmark$ & $\checkmark$ & 92.609 \\
            SK-101 & 448 & $\checkmark$ & $\checkmark$ & 93.712 \\
            Swin-Large & 224 &  & $\checkmark$ & 92.357 \\
            PVT-Large & 224 & $\checkmark$ & $\checkmark$ & 92.879 \\
            \hline
            \multicolumn{4}{l|}{Ensemble} & 94.750 \\
            \hline
		\end{tabular}
	}
	\vspace{-5pt}
	\caption{
    Partial results of the dynamic MLP in FGVC8~\cite{fgvc82021} on the iNaturalist 2021 dataset.
    ``Res'' denotes the resolution of the input image during training, and ``MS$_{\text{test}}$'' denotes multi-scale testing.
    \textbf{SK}: SK-Res2Net \cite{li2019selective,gao2019res2net}. \textbf{Swin}: Swin Transformer \cite{liu2021swin}. \textbf{PVT}: \cite{wang2021pyramid}. \textbf{Ensemble}: The ensemble result, including each model with an accuracy higher than 91.00\%.
	}
	\vspace{-5pt}
	\label{tab:ensemble}
\end{table}
\fi

%%%%%%%%%%%%%%%%%%%%%%%%%%%%%%%%%%%%%%%%%%%%%%%%%%%%%%%%%%%%%%%%%%%%%%%%%
% \vspace{-1pt}
\section{Experiment}
% \vspace{-1pt}
% first conduct the ablation study on iNaturalist 2021~\cite{van2021benchmarking}
% To evaluate the effectiveness of our dynamic MLP on fine-grained image classification, 
We conduct experiments on four fine-grained benchmark datasets that have collected additional information (iNaturalist 2017, 2018, 2021~\cite{van2018inaturalist,van2021benchmarking} and YFCC100M-GEO100~\cite{tang2015improving}).
% 检查一下是4个还是3个 ylf: inat2017 打算汇报与sota对比的结果
% which contain 579K | 437K | 2M | 87K training data.
% 95K | 24K | 100K | 2K
Notably, for a few images with unavailable metadata (corrupted or missing) in the dataset, we compensate with initialized zero input.
To evaluate the effectiveness of our dynamic MLP, we compare it with state-of-the-art fine-grained methods with and without utilizing additional information.
Furthermore, the ablation study and related analysis are conducted to better illustrate the network components of dynamic MLP.

\begin{table}[t]
	%\scriptsize
	\small
	\centering
	\renewcommand\arraystretch{1.1}
	\newcommand{\tabincell}[2]{\begin{tabular}{@{}#1@{}}#2\end{tabular}}
	%\begin{subtable}{0.44\textwidth}
	\resizebox{0.73\columnwidth}{!}{
		\begin{tabular}
            {p{0.64cm}<{\centering}|p{0.64cm}<{\centering}|cc|cc}
            \hline
            % \multicolumn{2}{c|}{\textbf{Dimension}} &
            % \multicolumn{2}{c|}{\textbf{iNat21 Mini}} & 
            % \multicolumn{2}{c}{\textbf{iNat21 Full}} \\
            % \hline
            % $d$ & $h$ & Top-1 & Top-5 & Top-1 & Top-5 \\
            $d$ & $h$ & 
            \multicolumn{2}{c|}{iNat21 Mini} & 
            \multicolumn{2}{c}{iNat21 Full} \\
            \hline\hline
            0 & -- & 76.102 & 91.456 & 86.178 & 96.400 \\
            \hline
            \multirow{4}{*}{256}
            % & 32 & 84.479 & 95.364 & 91.265 & 98.133 \\
            % & 64 & \textbf{84.694} & 95.337 & \textbf{91.397} & \textbf{98.213} \\
            % & 128 & 84.605 & \textbf{95.407} & 91.284 & 98.129 \\
            % & 256 & 84.566 & 95.337 & 91.382 & 98.181 \\
            % \hline
            % 64 & \multirow{4}{*}{64} & 84.324 & 95.188 & 91.258 & 98.095 \\
            % 128 & & 84.359 & 95.271 & 91.091 & 98.106 \\
            % 256 & & \textbf{84.694} & 95.337 & \textbf{91.397} & \textbf{98.213} \\
            % 512 & & 84.517 & \textbf{95.363} & 91.352 & 98.167 \\
            & 32 & 84.126 & 95.302 & 91.197 & 98.072 \\
            & 64 & \textbf{84.341} & 95.275 & \textbf{91.329} & \textbf{98.152} \\
            & 128 & 84.252 & \textbf{95.345} & 91.216 & 98.068 \\
            & 256 & 84.213 & 95.275 & 91.314 & 98.120 \\
            \hline
            64 & \multirow{4}{*}{64} & 83.971 & 95.126 & 91.190 & 98.034 \\
            128 & & 84.006 & 95.209 & 91.023 & 98.045 \\
            256 & & \textbf{84.341} & 95.275 & \textbf{91.329} & \textbf{98.152} \\
            512 & & 84.164 & \textbf{95.301} & 91.284 & 98.106 \\
            \hline
		\end{tabular}
	}
	\vspace{-5pt}
	\caption{
    % Performances of various $d$, $h$ in dynamic MLP-A with $N=2$. $d=0$ denotes the baseline (image-only).
    Top-1 (left) and top-5 (right) accuracy under various $d$ and $h$ with $N=2$, while $d=0$ denotes the baseline (image-only).
	}
	\vspace{-5pt}
	\label{tab:dimension}
\end{table}

\begin{table}[t]
	%\scriptsize
	\small
	\centering
	\renewcommand\arraystretch{1.1}
	\newcommand{\tabincell}[2]{\begin{tabular}{@{}#1@{}}#2\end{tabular}}
	%\begin{subtable}{0.44\textwidth}
	\resizebox{0.79\columnwidth}{!}{
		\begin{tabular}
            {c|cc|cc}
            \hline
            % & \multicolumn{2}{c|}{\textbf{iNat21 Mini}} & 
            % \multicolumn{2}{c}{\textbf{iNat21 Full}} \\
            % \hline
            % Stage Number ($N$) & Top-1 & Top-5 & Top-1 & Top-5 \\
            Stage Number ($N$) & 
            \multicolumn{2}{c|}{iNat21 Mini} & 
            \multicolumn{2}{c}{iNat21 Full} \\
            \hline\hline
            % 1 & 84.539 & \textbf{95.384} & 91.249 & 98.151 \\
            % 2 & \textbf{84.694} & 95.337 & \textbf{91.397} & \textbf{98.213} \\
            % 3 & 84.405 & 95.262 & 91.161 & 98.092 \\
            % 4 & 84.453 & 95.362 & 91.317 & 98.181 \\
            1 & 84.186 & \textbf{95.322} & 91.181 & 98.090 \\
            2 & \textbf{84.341} & 95.275 & \textbf{91.329} & \textbf{98.152} \\
            3 & 84.052 & 95.200 & 91.293 & 98.131 \\
            4 & 84.100 & 95.300 & 91.249 & 98.120 \\
            \hline
		\end{tabular}
	}
	\vspace{-5pt}
	\caption{
	Comparisons among different stage numbers.
	}
	\vspace{-5pt}
	\label{tab:num_stages}
\end{table}

\noindent\textbf{Implementation Details}:
During training, unless otherwise stated, the dynamic MLP and other existing works are both trained with the Stochastic Gradient Descent (SGD) algorithm based on the ResNet-50~\cite{he2016deep}, ResNet-101~\cite{he2016deep}, and SK-Res2Net-101~\cite{li2019selective,gao2019res2net} backbones for fair comparisons.
During inference, we apply a center crop to the image as the data augmentation.
More experimental details can be found in the Supplementary Materials.
% \\xxx xxx xxx
%dataset and training/test
% Our experiments are conducted on iNaturalist 2017, 2018, 2021~\cite{van2018inaturalist,van2021benchmarking} (we refer to iNat17, iNat18 and iNat21 below for brevity) and YFCC100M~\cite{thomee2016yfcc100m} benchmark with 
% More dataset details can be found in Supplementary Materials.
% At end we analyze the efficiency of our method.

%%%%%%%%%%%%%%%%%%%%%%%%%%%%%%%%%%%%%%%%%%%%%%%%%%%%%%%%%%%%%%%%%%%%%%%%%%%%
% 与SOTA的对比
%%%%%%%%%%%%%%%%%%%%%%%%%%%%%%%%%%%%%%%%%%%%%%%%%%%%%%%%%%%%%%%%%%%%%%%%%%%%
\subsection{Comparisons with State-of-the-arts}
Table~\ref{tab:sota} shows the top-1 and top-5 classification accuracy of our dynamic MLP (specifically refer to the dynamic MLP-C) and other approaches on various fine-grained datasets.
Specifically, we reproduce PriorsNet~\cite{mac2019presence} following their official descriptions\footnote{\link{https://github.com/macaodha/geo\_prior}} under unified backbones for a fair comparison. The hyperparameter in the post-processing method of GeoNet~\cite{chu2019geo} follows the settings in FGTL\footnote{\link{https://github.com/richardaecn/cvpr18-inaturalist-transfer}}~\cite{cui2018large}. More detailed settings of our re-implementation can be found in the Supplementary Materials.
Our dynamic MLP consistently achieves SOTA results on multiple datasets. On the iNaturalist 2021 dataset, we achieve a highly competitive 91.40\% top-1 accuracy using SK-Res2Net-101.
Additionally, it is noticed that the gap between our method and others is even more obvious for a bigger backbone model. For example, when changing the backbone from ResNet-50 to SK-Res2Net-101, the accuracy of ConcatNet~\cite{tang2015improving} increases by 5.38\% while ours is 5.94\% on the iNaturalist 2021 mini dataset.
Since dynamic MLP transforms image features under the guidance of multimodal features, we suspect that a bigger backbone can facilitate the dynamic MLP to explore stronger feature correlations and achieve more improvement compared to other methods.
% when the model is bigger, the gap between other methods and ours gets larger 
% and has higher accuracy when the models are bigger.
% 在各个数据集上的实验结果(baseline，其他论文方法，我们的方法)
% geonet的参数设置follow一个他们论文中的
Next, we compare our methods on the iNaturalist 2017 dataset with the best-published results from state-of-the-art fine-grained image-only methods~\cite{touvron2019FixRes,he2021transfg}.
Table~\ref{tab:inat17} shows that dynamic MLP outperforms existing works which even necessitate higher memory costs, more complex pipelines, and higher training or testing resolution.
% Existing multimodal methods are also compared.

In the iNaturalist 2021 challenge at FGVC8~\cite{fgvc82021}, we experiment with our dynamic MLP based on CNN and Transformer backbones with multiple image resolutions. During inference, we employ multi-scale testing and standard Ten-crop~\cite{krizhevsky2012imagenet} as post-processing strategies. The ensemble result achieves 94.75\% top-1 accuracy on the iNaturalist 2021 benchmark~\cite{fgvc82021}. 
% (see Table~\ref{tab:ensemble}). 这个表格就放补充材料了
See more detailed results in the Supplementary Materials.

\begin{table}[t]
	%\scriptsize
	\small
	\centering
    \renewcommand\arraystretch{1.1}
	\newcommand{\tabincell}[2]{\begin{tabular}{@{}#1@{}}#2\end{tabular}}
	%\begin{subtable}{0.44\textwidth}
	\resizebox{\columnwidth}{!}{
		\begin{tabular}
		%p{0.56cm}<{\centering}
            {
            % l|c|c|c|cc|cc
            % l|c|p{0.37cm}<{\centering}|p{0.37cm}<{\centering}|cc|cc
            l|p{0.44cm}<{\centering}|p{0.44cm}<{\centering}|p{0.44cm}<{\centering}|cc|cc
            % l|p{0.4cm}<{\centering}|p{0.4cm}<{\centering}|p{0.4cm}<{\centering}
            % |p{0.8cm}<{\centering}p{0.8cm}<{\centering}
            % |p{0.8cm}<{\centering}p{0.8cm}<{\centering}
            }
            \hline
            % \specialrule{0em}{0pt}{1.1pt}
            Method & Ver & IP & MP &
            \multicolumn{2}{c|}{iNat21 Mini} & 
            \multicolumn{2}{c}{iNat21 Full} \\
            \hline\hline
            % \specialrule{0em}{0pt}{1.1pt}
            Baseline & -- & -- & -- & 76.102 & 91.456 & 86.178 & 96.400 \\
            \hline
            % \specialrule{0em}{0pt}{1.1pt}
            \multirow{5}{*}{\tabincell{l}{Dynamic MLP}}
            & A & & & 84.341 & 95.275 & 91.329 & 98.152 \\
            % \hline
            % \specialrule{0em}{0pt}{1.1pt}
            & B & & & 84.406 & \textbf{95.403} & 91.244 & 98.118 \\
            % \hline
            % \specialrule{0em}{0pt}{1.1pt}
            & C & & \checkmark & 84.443 & 95.357 & 91.212 & 98.160 \\
            % \cline{2-7}
            % \specialrule{0em}{0pt}{1.1pt}
            & C & \checkmark & & 84.573 & 95.337 & 91.235 & 98.132 \\
            % \cline{2-7}
            % \specialrule{0em}{0pt}{1.1pt}
            & C & \checkmark & \checkmark & \textbf{84.694} & 95.337 & \textbf{91.397} & \textbf{98.213} \\
            \hline
		\end{tabular}
	}
	\vspace{-5pt}
	\caption{
	Comparisons between different dynamic MLP structures.
	``Ver'' denotes the version.
	``IP'' and ``MP'' indicate the input feature type for the image path and multimodal path, respectively.
	``\checkmark'' means the input is replaced with the concatenated feature. 
    % as input in replacement of the original input, i.e., the image feature or multimodal feature.
    %Dynamic MLP-C (a) and (b) respectively replace extra and image feature with concatenated feature.
    %(c) replaces both (Fig.~\ref{fig:dynamic_mlp} (c)).
	}
	\vspace{-6pt}
	\label{tab:structure}
\end{table}

\begin{table}[t]
	%\scriptsize
	\small
	\centering
	\renewcommand\arraystretch{1.1}
	\newcommand{\tabincell}[2]{\begin{tabular}{@{}#1@{}}#2\end{tabular}}
	%\begin{subtable}{0.44\textwidth}
	\resizebox{0.885\columnwidth}{!}{
		\begin{tabular}
            {l|c|c|cc}
            \hline
            % & \multicolumn{2}{c|}{\textbf{iNat21 Mini}} & 
            % \multicolumn{2}{c}{\textbf{iNat21 Full}} \\
            % \hline
            % Method & Top-1 & Top-5 & Top-1 & Top-5 \\
            % \hline\hline
            % (a) Image-only & 76.102 & 91.456 & 86.178 & 96.400 \\
            % \hline
            % (b) Concatenate & 83.706 & 94.932 & 91.200 & 97.992 \\
            % (c) Addition & 84.010 & 94.935 & 91.102 & 97.980 \\
            % (d) Multiply & 80.277 & 92.743 & 91.137 & 97.831 \\
            % \hline
            % (e) Dynamic MLP \textbf{(ours)} & \textbf{84.694} & \textbf{95.337} & \textbf{91.397} & \textbf{98.213} \\
            Method              & Flops & \#Params & Top-1 & Top-5 \\
            \hline\hline
            (a) Baseline        & 8.9 G & 66.8 M  & 76.102 & 91.456 \\
            \hline
            (b) Concatenation   & 8.9 G & 70.0 M  & 83.706 & 94.932 \\
            (b) Concatenation*  & 8.9 G & 74.2 M  & 83.915 & 94.893 \\
            \hline
            (c) Addition        & 8.9 G & 69.9 M  & 84.010 & 94.935 \\
            (c) Addition*       & 8.9 G & 74.2 M  & 84.068 & 94.982 \\
            \hline
            (d) Multiplication  & 8.9 G & 69.9 M  & 80.277 & 92.743 \\
            (d) Multiplication* & 8.9 G & 74.2 M  & 81.649 & 93.421 \\
            \hline
            (e) Dynamic MLP \textbf{(ours)} 
                                & 8.9 G & 74.2 M & \textbf{84.694} & \textbf{95.337} \\ 
            \hline
		\end{tabular}
	}
	\vspace{-5pt}
	\caption{
    %classification performance of 
	Comparisons of our dynamic MLP and existing multimodal fusion methods on the iNaturalist 2021 mini dataset. (a): Baseline. (b): Concatenation. (c): Addition. (d): Multiplication. (e): Dynamic MLP.
	*Compensated models with the same flops and parameter numbers as our dynamic MLP.
	}
	\vspace{-5pt}
	\label{tab:existing_works}
\end{table}

% \begin{figure}[t]
% 	\vspace{0pt}
% 	\begin{center}
% 		\setlength{\fboxrule}{0pt}
% 		\vbox{\includegraphics[width=\columnwidth]{./figs/tsne_feature.pdf}}
% 	\end{center}	
% 	\vspace{-10pt}
% 	\caption{
% 	 Comparisons of image representations between Turdus merula and Turdus fuscater with and without dynamic MLP, where our method generates more discriminative representations and achieves higher recognition accuracy.
% 	}
% 	\label{fig:tsne_feature}
% 	\vspace{-8pt}
% \end{figure}

\subsection{Ablation Study}\label{sec:ablation_study}
%%%%%%%%%%%%%%%%%%%%%%%%%%%%%%%%%%%%%%%%%
% 第一个对比试验：通道数/特征维度数
%%%%%%%%%%%%%%%%%%%%%%%%%%%%%%%%%%%%%%%%%
\noindent\textbf{Channel Dimension (i.e., $d$, $h$)}: 
We first examine the impact of input and hidden channel dimensions (Fig.~\ref{fig:stage_number}) on dynamic MLP-A on classification performance.
Specifically, we explore their effect by fixing one and adjusting the other in Table~\ref{tab:dimension}. 
The experiments are conducted based on the SK-Res2Net-101 backbone on the iNaturalist 2021 dataset.
It is observed that $d=256$, $h=64$ steadily achieve higher accuracy compared to other dimension combinations.
We assume that a higher dimension may lead to overfitting of the network, while $d=256$, $h=64$ are sufficient to derive considerable discriminative representations.

%%%%%%%%%%%%%%%%%%%%%%%%%%%%%%%%%%%%%%%%%
% 第二个对比试验：层数/循环次数/stage number/layer number
%%%%%%%%%%%%%%%%%%%%%%%%%%%%%%%%%%%%%%%%%
\noindent\textbf{Stage Number (i.e., $N$)}:
We then explore the best stage number for dynamic MLP-A by setting it under $d=256$ and $h=64$.
%performs the best and
Table~\ref{tab:num_stages} shows that $N=2$ achieves the best performance trade-off, avoiding the overfitting risks with more stages applied.
Therefore, we default to using $N=2$ in the following experiments.
% This fusion process is iterated until the latest feature map S1 is generated
% \rewrite{Since the first stage of dynamic MLP introduces the additional information to the image representations initially, the output essentially carries the discriminative contextual clues.
% We suspect that the latter stages mainly focus on adjusting the representations precisely, thus $N=2$ is adequate for our architecture. }

%%%%%%%%%%%%%%%%%%%%%%%%%%%%%%%%%%%%%%%%%
% 第三个对比试验：图像特征以及额外特征具体是采用concat的之后的还是它们本身。
%%%%%%%%%%%%%%%%%%%%%%%%%%%%%%%%%%%%%%%%%

\noindent\textbf{Comparison of Structures}:
In this section, we compare the dynamic-MLP-A with its two variants (Fig.~\ref{fig:dynamic_mlp}) based on the optimal settings ($d=256$, $h=64$, $N=2$).
% Notably, we explore three types of dynamic MLP-C according to different input sources for the image path and multimodal path:
% (\romannumeral1): image feature and concatenated feature
% (\romannumeral2): Take as input the concatenated feature and multimodal feature, respectively.
% (\romannumeral3): Take the concatenated feature as inputs for both paths.
% denote the architecture that replace the input multimodal feature, image feature and both with their concatenation, respectively.
Table~\ref{tab:structure} indicates that the dynamic MLP-C achieves superior performance on fine-grained image classification, which is in line with our expectations that a deeper MLP projection and a concatenated operation before feature fusion can further benefit our architecture.
% Since dynamic MLP-B only deepens the structure and dynamic MLP-C further utilizes the mutual feature of both paths as input to benefit the classification.
% 我们的方法都远超image-only的方法，并且随着改进，性能不断的有提升。
% 但是可能有人会问，我们的方法增加了模型参数量，那么会不会其他方法增加参数量之后也可以变好呢？
% 我们做了实验，说明即使给其他方法添加额外的fc层，他们也仅有微弱涨点，还是不如我们，详情见附录

\begin{figure}[t]
	\vspace{0pt}
	\begin{center}
		\setlength{\fboxrule}{0pt}
		\vbox{\includegraphics[width=\columnwidth]{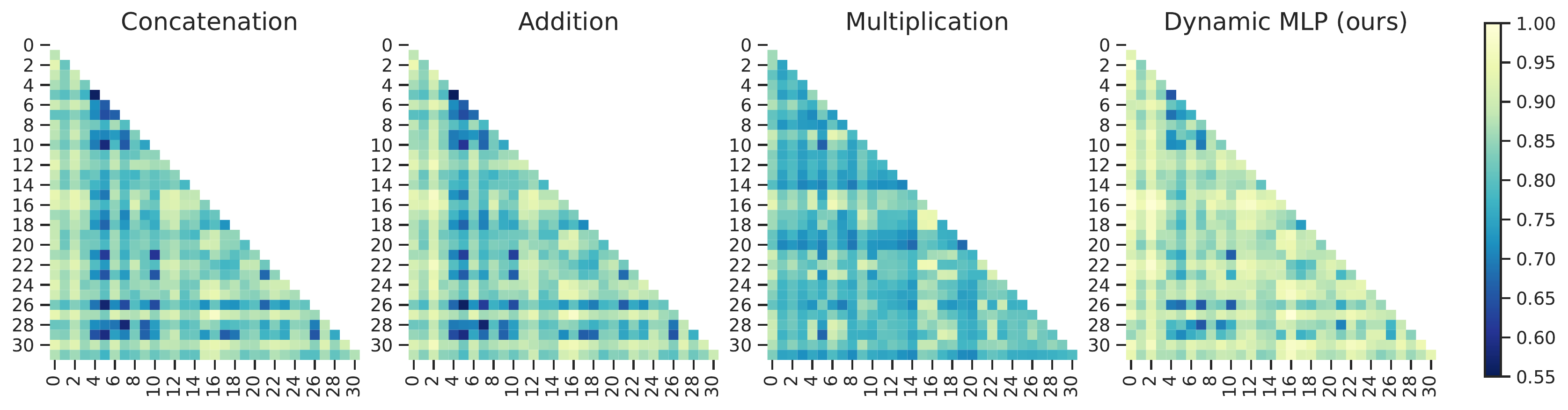}}
	\end{center}	
	\vspace{-10pt}
	\caption{
    Visualization of the L2 distance between each two feature representations of sampled categories. A lighter color denotes a farther distance and better distinction.
    }
	\label{fig_quantitative_measure_of_clusters}
	\vspace{-5pt}
\end{figure}

%%%%%%%%%%%%%%%%%%%%%%%%%%%%%%%%%%%%%%%%%
% 第四个对比试验：与图2中提到的(a)Image-only (b)Concatenate (c)Addition (d)Multiply (e)Ours对比
%%%%%%%%%%%%%%%%%%%%%%%%%%%%%%%%%%%%%%%%%
\noindent\textbf{Comparison of Fusion Strategies}: 
% We speculate that there is more effective and efficient way to fuse multimodal futures.
% fix the input and hidden channel dimension and stage number of our dynamic MLP-C (e) ($d =256$, $h = 64$, $N = 2$)
% Therefore our dynamic MLP can be more effective and efficient than former works.
To verify the effectiveness of our method, we compare our dynamic MLP (e) with the baseline (a) and various existing fusion strategies, i.e., concatenation (b), addition (c), and multiplication (d) (Fig.~\ref{fig:existing_works}).
Specifically, we use dynamic MLP-C and fix its channel dimension and stage number to the optimal settings ($d=256$, $h=64$, $N=2$) for all experiments on the iNaturalist 2021 mini dataset.
Notably, our model has higher flops and parameter numbers. 
For a fair comparison, we compensate the networks of previous methods with additional MLP layers. The adjusted models have the same flops and parameters as ours.
% Table~\ref{tab:existing_works} demonstrates that our dynamic MLP outperforms previous works provided the same model capacity.
As shown in Table~\ref{tab:existing_works}, our dynamic MLP outperforms previous work under the unified model capacity setting.
%All of the existing multimodal fusion strategies neglect the feasible adjustment on image representation according to the metadata. To the best of our knowledge, our proposed dynamic MLP is the first to utilize conditional parameters generated from geographical and temporal information to project the image representations.
% \\xxx xxx xxx

%%%%%%%%%%%%%%%%%%%%%%%%%%%%%%%%%%%%%%%%%%%%%%%%%%%%%%%%%%%%%%%%%%%%%%%%%%%%
\subsection{Analysis}
Although the dynamic MLP is demonstrated to improve the performance across multiple fine-grained datasets, we would like to understand how its mechanism operates. More analyses can be found in the Supplementary Materials.
% \\xxx xxx xxx

%%%%%%%%%%%%%%%%%%%%%%%%%%%%%%%%%%%%%
% 拉开了特征之间的距离
%%%%%%%%%%%%%%%%%%%%%%%%%%%%%%%%%%%%%
\noindent\textbf{Visualization of T-SNE Representations}: 
We evaluate 16 species from the genus Turdus in the iNaturalist 2021 dataset and visualize their image representations as points based on well-trained models of different methods, respectively.
Fig.~\ref{fig:tsne} shows that our dynamic MLP produces more discriminative image representations than previous works.
As illustrated in Sec.\ref{sec:interact_dimensions}, former works fuse image and location features into one dimension, which is insufficient for separating the mutual representations. It is also shown in Fig.~\ref{fig:tsne} (b) that the concatenation strategy stretches the cluster distribution only in one direction.
% Our motivation is that different species with the same appearance have similar image representations derived by a unified network, which is hard to be distinguished.
Since dynamic MLP transforms the image representation by weighted features generated from the additional information, it involves mutual interaction on a broader dimension. Therefore, it improves the representation ability of the feature to benefit fine-grained image classification.
% To verify its effectiveness, 
% To verify, we evaluate two species depicted in Fig.~\ref{fig:geographical_temporal}, i.e., Turdus merula and Turdus fuscater using the image-only method and our dynamic MLP, respectively. 
% Then we visualize the relative distribution of their features by t-SNE~\cite{van2008visualizing} in Fig.~\ref{fig:tsne_feature}, where features refined by our method are more discriminative, leading to better performance (9.67\% and 16.34\% improvement for Turdus merula and Turdus fuscater, respectively).
% image-only 'acc1:68.1667', 'acc5:84.6667', 'acc1_4284:62.0000', 'acc5_4284:77.0000', 'acc1_4279:74.3333', 'acc5_4279:92.3333'
% dynamic MLP 'acc1:81.1667', 'acc5:91.5000', 'acc1_4284:71.6667', 'acc5_4284:84.6667', 'acc1_4279:90.6667', 'acc5_4279:98.3333'
% 来一张图，就是表示一下原本图像的feature挨得很近，但是经过MLP之后很远。

\noindent\textbf{Visualization of L2 Distance Between Image Representations}: 
% We calculate the L2 distance between class-wise representations based on weight of the classifier.the output of the classifier across the cluster center several groups of similar categories as a quantitative measurement of the image representations.
Each row of the classifier weight is a representative feature corresponding to a specific category. 
% Hence, their correlation reflects the model performance. 
To measure the instance representations quantitatively, we calculate the L2 distance between every two of the class-wise representative features
% randomly sampled class vectors 
and produce the heatmap visualization. Fig.~\ref{fig_quantitative_measure_of_clusters} shows that the dynamic MLP can have more discriminant feature representations.

% \implusrewrite{check:} % A2 答非Q2所问。A2的Q应该是 t-SNE details for the concatenation methods
% 主要解决所有关于t-SNE的问题，包括除了t-SNE以外，用一个量化的指标来评价特征的区分程度，目前考虑用L2 distance between logit
% 这边可以做一个棋盘格子的热度图，距离越远，数字越大，颜色越深，可以并排着放3个棋盘格
% https://seaborn.pydata.org/generated/seaborn.heatmap.html

%%%%%%%%%%%%%%%%%%%%%%%%%%%%%%%%%%%%%
% 速度
%%%%%%%%%%%%%%%%%%%%%%%%%%%%%%%%%%%%%
\noindent\textbf{Training/Inference Efficiency}: 
% 绘制一个表，是training hours和testing FPS的，因为有一些方法是两阶段调整的。
Table~\ref{tab:speed} shows the training and inference efficiency of existing multimodal methods and our dynamic MLP on the iNaturalist 2021 mini dataset.
Specifically, GeoNet~\cite{chu2019geo} is extremely time-consuming as its framework requires fine-tuning on a pretrained image-only model.
Notably, our proposed dynamic MLP achieves top performance (84.694\%) while still maintaining competitive training and inference efficiency.
% \\xxx xxx xxx

\begin{table}[t]
	%\scriptsize
	\small
	\centering
	\renewcommand\arraystretch{1.1}
	\newcommand{\tabincell}[2]{\begin{tabular}{@{}#1@{}}#2\end{tabular}}
	%\begin{subtable}{0.44\textwidth}
	\resizebox{0.68\columnwidth}{!}{
		\begin{tabular}
            {l|c|c|c}
            \hline
            Method & Acc (\%) & H $\downarrow$ & FPS $\uparrow$ \\
            \hline\hline
            Baseline & 76.102 & 30.0 & 19.3 \\
            \hline
            ConcatNet~\cite{tang2015improving} & 83.706 & 32.5 & 18.4 \\
            PriorsNet~\cite{mac2019presence} & 83.600 & 31.5 & 19.3 \\
            GeoNet~\cite{chu2019geo} & 81.360 & 55.0 & 19.0 \\
            EnsembleNet~\cite{terry2020thinking} & 80.277 & 31.5 & 18.4 \\
            Dynamic MLP \textbf{(ours)} & \textbf{84.694} & 32.0 & 18.4 \\
            \hline
		\end{tabular}
	}
	\vspace{-3pt}
	\caption{
	Comparisons of training and inference efficiency.
	``H'' denotes the total training hours for 90 epochs on iNaturalist 2021 mini dataset evaluated on 8 TITAN Xp GPUs.
	``FPS'' denotes the inference FPS measured on the same machine.
	}
	\vspace{-5pt}
	\label{tab:speed}
\end{table}
%%%%%%%%%%%%%%%%%%%%%%%%%%%%%%%%%%%%%%%%%%%%%%%%%%%%%%%%%%%%%%%%%%%%%%%%%
\section{Conclusion}
In this paper, we propose an end-to-end trainable framework, i.e., the dynamic MLP, to improve the effectiveness of geographical and temporal information on fine-grained image classification. Our method adaptively projects image representations by weights conditioned on additional information, which involves high-dimensional feature interaction. To our best knowledge, we are the first to introduce dynamic MLP in multimodal fine-grained image classification. Notably, our method achieves SOTA results on multiple fine-grained datasets. Further, the t-SNE visualization verifies that its image representations are more discriminative than previous works. We hope dynamic MLP can serve as a lightweight yet effective baseline for multimodal fine-grained image classification.

% We introduce a spatio-temporal prior to help disambiguate fine-grained categories resulting in improved test time image classification performance. In addition to helping image classification, our model also naturally captures
% the relationships between locations and objects, objects and
% objects, photographers and objects, and photographers and
% locations in an interpretable manner. Importantly, our prior
% is efficient at test time, both in terms of model size and inference speed, and scales to large numbers of categories.

% We have given a systematic overview of geo-aware networks for fine-grained recognition. To deal with the lack of
% fine-grained geolocation datasets, we introduced the iNaturalist and YFCC100M fine-grained geolocation datasets.
% Experimental results show that all geo-aware networks
% achieve significant improvements over image-only models.
% Specifically, the post-processing model performs best on
% large baseline models, while the feature modulation model
% performs best on small baseline models and even outperforms the large image-only model. Although experiments
% in this paper are mainly on animal and plant species recognition, we believe that the geo-aware networks examined
% in this paper are generally useful and can be easily extended for recognizing any location sensitive fine-grained
% categories, such as car’s make/model and food.

%%%%%%%%%%%%%%%%%%%%%%%%%%%%%%%%%%%%%%%%%%%%%%%%%%%%%%%%%%%%%%%%%%%%%%%%%
\noindent\textbf{Acknowledgement.}  
% The authors would like to thank the editor and the anonymous reviewers for their critical and constructive comments and suggestions. 
This work was supported by
Postdoctoral Innovative Talent Support Program of China under Grant BX20200168, 2020M681608, %xiangli
and National Science Fund of China under Grant No. U1713208.
% , Program for Changjiang Scholars, and “111” Program AH92005.%jianyang

%%%%%%%%%%%%%%%%%%%%%%%%%%%%%%%%%%%%%%%%%%%%%%%%%%%%%%%%%%%%%%%%%%%%%%%%%
% 一些数据集
% Brian L Sullivan, Christopher L Wood, Marshall J Iliff,
% Rick E Bonney, Daniel Fink, and Steve Kelling. ebird: A
% citizen-based bird observation network in the biological sciences. Biological Conservation, 2009.

% GBIF - www.gbif.org. 2019.

% iNaturalist - www.inaturalist.org. 2019.

%%%%%%%%%%%%%%%%%%%%%%%%%%%%%%%%%%%%%%%%%%%%%%%%%%%%%%%%%%%%%%%%%%%%%%%%%

%%%%%%%%% REFERENCES
{\small
\bibliographystyle{ieee_fullname}
\bibliography{egbib}
}

\end{document}